\pdfoutput=1

\documentclass[11pt]{article}

\usepackage{EMNLP2022}

\usepackage{times}
\usepackage{latexsym}
\usepackage{booktabs}

\usepackage[T1]{fontenc}

\usepackage[utf8]{inputenc}

\usepackage{microtype}

\usepackage{inconsolata}

\usepackage{graphicx}
\usepackage{amsmath}

\usepackage{caption}
\usepackage{subfig,caption}
\usepackage{enumitem}

\usepackage{xcolor}

\newcommand{\name}{ConCoRD}
\newcommand{\fullnameacronym}{Consistency Correction through Relation Detection}

\newcommand{\fullnametitle}{Consistency Correction through Relation Detection}

\DeclareMathOperator*{\argmax}{argmax}

\definecolor{mygreen}{RGB}{50, 128, 50}
\newcommand{\Sylvia}[1]{}
\newcommand{\quotestring}[1]{\textit{#1}}

\newif\ifcomments
\commentstrue
\ifcomments
\usepackage[normalem]{ulem}
\definecolor{CMpurple}{rgb}{0.6,0.18,0.64}
\newcommand\cm[1]{\textcolor{CMpurple}{\textsf{\scriptsize[\textbf{CM\@:} #1]}}}
\newcommand\cmi[1]{\textcolor{CMpurple}{#1}}
\newcommand\cmm[1]{\marginpar{\raggedright\tiny\textcolor{CMpurple}{\textsf{{\bfseries CM\@:} #1}}}}
\newcommand\cms{\bgroup\markoverwith{\textcolor{CMpurple}{\rule[.4ex]{2pt}{0.8pt}}}\ULon}
\else
\newcommand\cm[1]{}
\newcommand\cmi[1]{\ignorespaces}
\newcommand\cmm[1]{}
\newcommand\cms[1]{#1}
\fi

\title{Enhancing Self-Consistency and Performance of Pre-Trained Language Models through Natural Language Inference}

\author{Eric Mitchell, Joseph J. Noh, Siyan Li, William S.~Armstrong, \\
{\bf Ananth Agarwal, Patrick Liu, Chelsea Finn, Christopher D. Manning} \\
Stanford University \\
\texttt{\href{mailto:eric.mitchell@cs.stanford.edu}{eric.mitchell@cs.stanford.edu}}}

\begin{document}
\maketitle
\begin{abstract}


While large pre-trained language models are powerful, their predictions often lack logical consistency across test inputs. For example, a state-of-the-art Macaw question-answering (QA) model answers \textit{Yes} to \textit{Is a sparrow a bird?} and \textit{Does a bird have feet?} but answers \textit{No} to \textit{Does a sparrow have feet?}. To address this failure mode, we propose a framework, {\fullnametitle}, or \textbf{\name}, for boosting the consistency and accuracy of pre-trained NLP models using pre-trained natural language inference (NLI) models without fine-tuning or re-training. Given a batch of test inputs, {\name} samples several candidate outputs for each input and instantiates a factor graph that accounts for both the model's belief about the likelihood of each answer choice in isolation and the NLI model's beliefs about pair-wise answer choice compatibility. We show that a weighted MaxSAT solver can efficiently compute high-quality answer choices under this factor graph, improving over the raw model's predictions. Our experiments demonstrate that {\name} consistently boosts accuracy and consistency of off-the-shelf closed-book QA and VQA models using off-the-shelf NLI models, notably increasing accuracy of LXMERT on ConVQA by 5\% absolute. See the project website\footnote{\href{https://ericmitchell.ai/emnlp-2022-concord/}{\url{https://ericmitchell.ai/emnlp-2022-concord/}}} for code and data.
\end{abstract}

\section{Introduction}

\begin{figure}
    \centering
    \includegraphics[width=0.82\columnwidth]{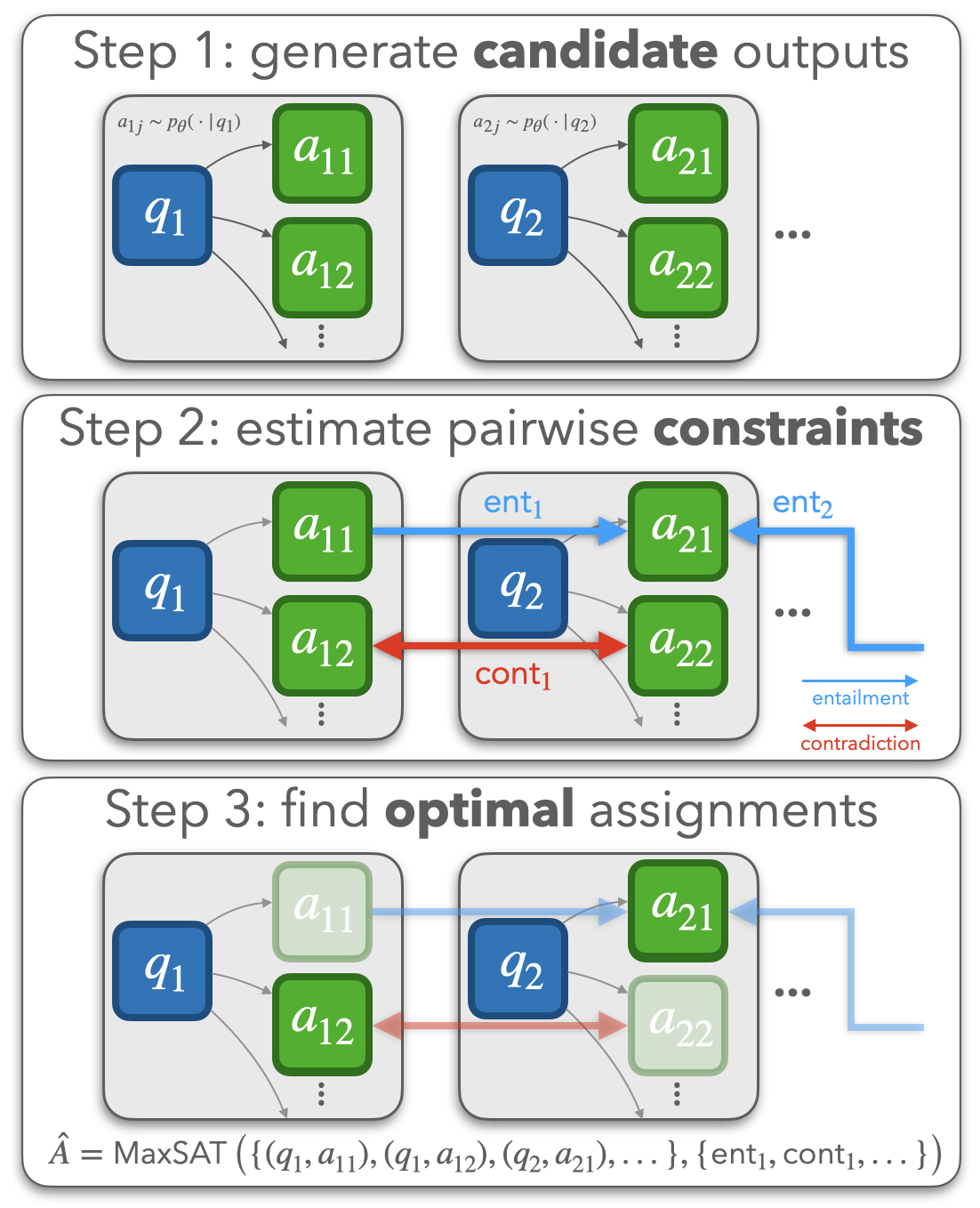}
    \vspace{-2mm}
    \caption{\footnotesize {\name} first generates candidate outputs from the base pre-trained model, then estimates soft pairwise constraints between output choices, and finally finds the most satisfactory choices of answers accounting for both the base model and NLI model's beliefs.}
    \vspace{-5mm}
    \label{fig:method-overview}
\end{figure}

Reliable and trustworthy AI systems should demonstrate internal \textit{self-consistency}, in the sense that their predictions across inputs should imply logically compatible beliefs about the world. However, even powerful large language models are known to lack self-consistency \citep{ray-etal-2019-sunny,elazar2021measuring,kassner2021beliefbank}.
For example, a question-answering (QA) model that answers the question \textit{Is a sparrow a bird?} and \textit{Does a bird have feet?} with \textit{Yes} is implicitly expressing the belief that \textit{A sparrow is a bird} and \textit{A bird has feet}. If the same model answers the question \textit{Does a sparrow have feet?} with \textit{No}, the model expresses the logically incompatible belief \textit{A sparrow does not have feet}. In such cases, ascertaining the model's ``true'' belief is difficult, making interpreting and validating its behavior correspondingly challenging.

Prior work has improved model self-consistency by training with specialized loss functions \citep{elazar2021measuring} or data augmentation \citep{ray-etal-2019-sunny}, or alternatively re-ranking model predictions based on their mutual self-consistency using pre-written logical constraints, such as ``all mammals have fur'' \citep{kassner2021beliefbank}. However, the first class of methods requires expensive fine-tuning which might be impractical for many practitioners for very large pre-trained models, and re-ranking methods require an explicit collection of the logical relations of interest, making scaling a challenge. Still, re-ranking-based approaches have the benefit of not requiring fine-tuning, and we hypothesize that their scalability limitations may be addressed by \textit{estimating} logical relationships between model predictions on the fly. Specifically, we hypothesize that existing pre-trained natural language inference (NLI) models can estimate logical relationships between an arbitrary pair of model predictions well enough to provide an effective, scalable substitute for explicit collection of such constraints. Leveraging these estimated constraints, we can construct a factor graph representing a probability distribution over model outputs that incorporates both the original model's confidence scores and the NLI model's beliefs about logical relationships.

Our primary contribution is {\fullnameacronym}, or \textit{\name}, a framework to improve the consistency and performance of a pre-trained \textit{base language model} without fine-tuning by using more confident and better attested model predictions to override less confident model beliefs. See Figure~\ref{fig:method-overview} for an overview. To enable propagation of model beliefs, we estimate pair-wise logical relationships between model predictions using a pre-trained NLI model. Using these pair-wise relationships, we define an undirected graphical model representing a distribution over responses accounting for both the base model's beliefs and the NLI model's estimates of answer compatibility. We efficiently find the approximate mode of this distribution among the base model's top answer choices for each input as the solution of a MaxSAT problem, which consistently produces more accurate and self-consistent predictions than using the raw model predictions. In Section~\ref{sec:beliefbank} we find that {\name} produces an 8.1\% absolute improvement in F1 of a pre-trained Macaw model \citep{tafjord2021general} on the BeliefBank QA dataset \citep{kassner2021beliefbank}. In Section~\ref{sec:cvqa} we find a 5.0\% absolute improvement in accuracy of a pre-trained LXMERT model \citep{tan2019lxmert} on the ConVQA dataset \citep{ray-etal-2019-sunny}, and in Section~\ref{sec:editing} we find that {\name} enables test-time \textit{model editing} \citep{Sinitsin2020Editable,mitchell2021fast}, updating model predictions at test time when presented with new information.

\section{Related Work}



Prior work for maintaining consistency in the question-answering space often involves additional training to improve performance. \citet{chen2021can} transform the Natural Questions \citep{nq} dataset question-answer pairs into premise-hypothesis pairs, then uses an NLI model trained on this dataset as a decider for unanswerable questions. \citet{alberti2019synthetic} generate questions from unlabeled texts, then filter them to ensure roundtrip consistency; pre-training on this synthetic set improves performance on SQuAD 2.0 \citep{squad2} and Natural Questions. \citet{asai2020logic} augment QA-pairs with their logically symmetric and transitive counterparts through linguistic approaches to enhance cross-dataset QA performance. {\name} differs significantly from these question-answering-specific approaches because no fine-tuning of the base model is needed and the methodology is not specific to question-answering. 

\begin{figure*}
    \centering
    \includegraphics[width=\textwidth]{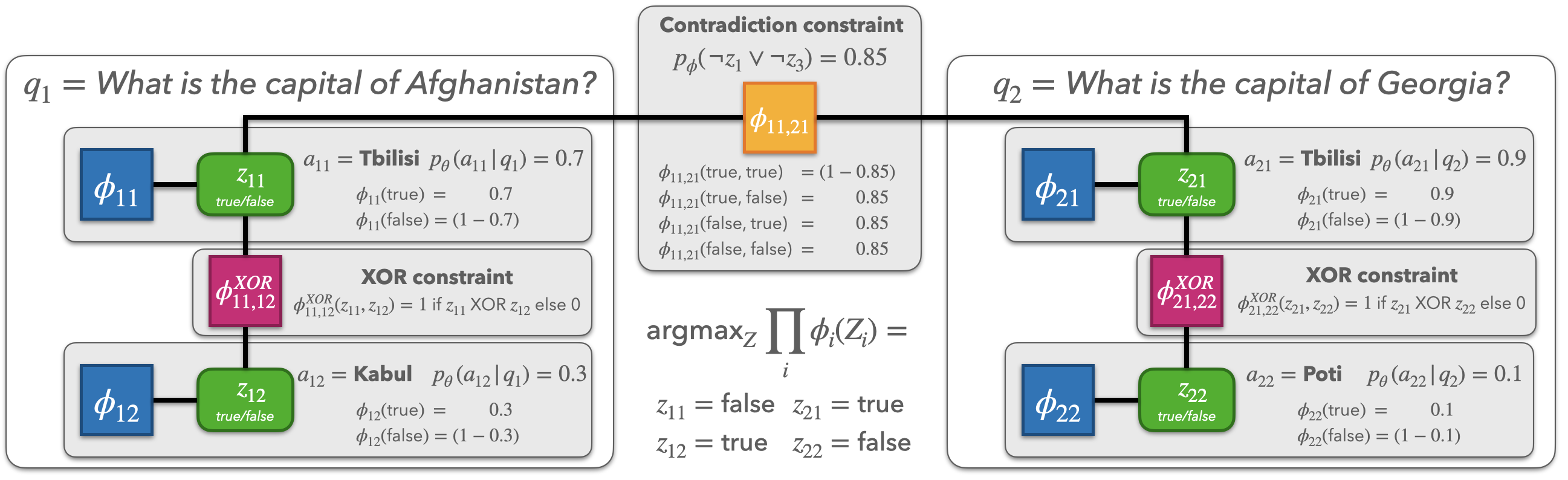}
    \caption{\footnotesize An example factor graph for a simplified batch with two questions, $q_1$ = \textit{What is the capital of Afghanistan?} and $q_2$ = \textit{What is the capital of Georgia?}. Although \textit{Tbilisi} is the most likely answer for both questions, the assignment of variables that is best under the estimated contradiction constraint flips the answer to the first question to \textit{Kabul}. The top-2 answer choices for each question are sampled from the base model, and a soft contradiction constraint is detected between variables $z_1$ (representing the truth of the answer \textit{Tbilisi} for $q_1$) and $z_3$ (representing the truth of the answer \textit{Tbilisi} for $q_2$).}
    \vspace{-2mm}
    \label{fig:factor-graph}
\end{figure*}

Similarly to {\name}, \citet{kassner2021beliefbank} re-rank model predictions by solving an optimization problem defined by a combination of the base model confidence scores and pair-wise constraints representing the logical compatibility of different model predictions stored in a persistent memory, which they call BeliefBank. The key distinguishing property of {\name} is the fact that pair-wise constraints between model predictions are dynamically estimated by a pre-trained NLI model, rather than drawn from a fixed, pre-collected set of constraints. Dynamically estimating the constraints has a variety of benefits, eliminating the need for manually collecting the logical constraints of interest, automating the process of determining whether a particular constraint applies to a particular pair of predictions, and likely inheriting improvements in Natural language inference (NLI, \citet{maccartney-manning-2008-modeling}) models over time. 

NLI has long been used to maintain logical consistency in generated dialogue utterances \citep{welleck2018dialogue,dziri2019evaluating,song2020generating}, radiology report domain entities \cite{miura2020improving}, and summarization \citep{laban2021summac,honovich2022true}. Perhaps most similarly, \citet{jung2022maieutic} use NLI to estimate constraints between factual statements produced by GPT-3. These prior approaches support our intuition for using NLI models to improve logical consistency among batches of answers. While the authors explore applications of this framework to multi-step reasoning for True/False questions or statements, our work focuses on applying this methodology to more general settings, such as VQA, open-ended QA, and model editing.


\section{{\fullnametitle}}

{\name} contains three key components, the \textit{base model}, a \textit{relation model} (typically a pre-trained NLI model), and an \textit{inference procedure} that combines the predictions of the two models into a more accurate and self-consistent set of beliefs. Importantly, both the base model and relation model are pre-trained, off-the-shelf models; {\name} does not update any weights or require training data for either model,
using only a small validation set for hyperparameter tuning. We next explain the function of each of these components when executing {\name}.

\subsection{Base Model}
The core function of the base model in {\name} is generating a set of \textit{candidate outputs} for a given input, which are ultimately re-ranked by the inference process (Sec.~\ref{sec:inference}).
Given a batch of $N$ model queries $Q=\{q_i\}$, the first step of {\name} is to generate a set of $J$ candidate outputs for each query $\hat A_i = \{\hat a_{i1},...,\hat a_{iJ}\}$,
along with their corresponding likelihoods $p_\theta(\hat{a}_{ij}|q_i)$. Note that the candidate outputs need not be an IID sample from the base model; for example, we might use beam search with a diversity bonus
to produce a more diverse set of candidates \citep{Vijayakumar2018diverse}. Each pair of query and candidate output forms a \textit{model belief} $b_{ij} = (q_i, \hat a_{ij})$; the output of the base model is the complete set of model beliefs $B = \{b_{ij}\}$ and their corresponding \textit{normalized} probabilities $p_\theta^{ij}$\footnote{Normalized such that $\sum_j p_\theta^{ij} = 1$.}. The base models in our experiments are pre-trained question-answering models based on T5-large \citep{2020t5} and pre-trained visual question-answering models such as LXMERT \citep{tan2019lxmert} and ViLT \citep{kim2021vilt}.

\subsection{Relation Model}
The relation model $p_\phi(\cdot | x, x')$ estimates the most likely logical relationship between an ordered pair of natural language utterances from the choices $\{\texttt{none},\,\texttt{fwd-entail},\,\texttt{contradict},$ $ \texttt{equivalence}\}$.\footnote{Because relationships are estimated between ordered pairs of utterances, we can form an equivalence relation if \texttt{fwd-entail} is predicted for both orderings of the utterances.} In addition to the model beliefs $B$, we define optional \textit{context statements} $c_{ijk} = C(b_{ij})$, $K$ relevant statements that may be retrieved, generated, or manually written for each model belief. The ability to incorporate context statements enables {\name} to modulate model behavior independently for each input in the test batch, rather than reasoning transductively about pairs of test inputs. See Table~\ref{tab:editing} for examples of model beliefs and context statements. Inputs to the relation model are either pairs of two model beliefs $(b_{ij}, b_{i'j'})$ or pairs of one model belief and one context statement $(b_{ij}, c_{ijk})$. We define the most likely inter-belief relation as $r_{ij,i'j'} = \argmax_r p_\phi(r|b_{ij}, b_{i'j'})$, and similarly for belief-context relations $r_{ijk} = \argmax_r p_\phi(r|b_{ij}, c_{ijk})$. The output of the relation model is the set of most-likely relations $R = \{r_{ij,i'j'}\} \cup \{r_{ijk}\}$ and their associated probabilities, which we denote as $p_\phi^{ij,i'j'}$ and $p_\phi^{ijk}$. Our experiments use various pre-trained NLI models based on RoBERTa \citep{liu2019roberta} and ALBERT \citep{lan2020albert} as the relation model.

\paragraph{Question-answer to statement conversion.}
While concatenating query $q_i$ and candidate output $\hat a_{ij}$ to produce inputs to the relation model is perhaps the simplest approach to estimating soft constraints, we use a statement conversion model to provide inputs to the relation model that are closer to its training distribution. Instead of defining the belief $b_{ij} = (q_i, \hat a_{ij})$ as concatenation of $q_i$ and $\hat a_{ij}$, we define $b_{ij}$ to be the statement $f_\psi(q_i, \hat a_{ij})$, where $f_\psi$ is the conversion model. We fine-tune a small T5 model on a combination of data from \citep{demszky2018transforming} and BeliefBank \citep{kassner2021beliefbank} to produce a model that maps a (question, answer) pair into a natural language statement. Details about the fine-tuning procedure and data are provided in Appendix~\ref{sec:qaconv-appendix}.

\subsection{Inference}
\label{sec:inference}
{\name}'s inference procedure maps the set of beliefs $B$ and pair-wise relations $R$ into a choice of the most likely belief for each question. To define the inference problem, we first define a binary decision variable $z_{ij}$ representing the estimated truth value of model belief $b_{ij}$. A value of 1 for node $z_{ij}$ in the maximum likelihood configuration means that $\hat{a}_{ij}$ is returned for query $q_i$; the problem includes a constraint that \textit{exactly} one candidate answer is true for each query.
The factor graph includes the set of variables $Z = \{z_{ij}\}_{i,j=1,1}^{N,J}$ and various factors (functions mapping a subset of $Z$ to a non-negative scalar) derived from the base model and relation model's beliefs and the hard constraint of returning only one answer per question. Factors are defined such that more desirable configurations of $z_{ij}$ yield a larger \textit{product} of the individual factors. First, unary factors $\phi_{ij}(z_{ij})$ encode the base model's beliefs about the likelihood of specific answers, and are defined as:
\begin{equation}
    \phi_{ij}(z_{ij}) = \begin{cases}
    \frac{p_{ij}}{1-p_{ij}} & \text{if}\;z_{ij} = 1 \\
    1 & \text{otherwise}\end{cases}
\end{equation}
where $p_{ij} = p_\theta(\hat{a}_{ij}|q_i)$; in other words, the factor takes the odds ratio if the corresponding statement variable $z_{ij}$ is assigned a truth value of 1; otherwise, the factor takes value 1. In order to encode the hard constraint that exactly one output should be returned for each query, we include a $J$-ary factor $\phi_i(Z_i)$ for each group of nodes $Z_i = \{z_{ij}\}_{j=1}^{J}$, which is equal to 1 for configurations where exactly one of the nodes takes a value of 1, and 0 for all other configurations.

Binary factors $\phi_{ij,i'j'}(z_{ij}, z_{i'j'})$ and optionally $\phi_{ijk}(z_{ij}, c_{ijk})$ encode compatibility between pairs of model beliefs (or model belief-context pairs):
\begin{equation*}
    \phi_{ij,i'j'}(z_{ij}, z_{i'j'}) = \begin{cases}1 & \hspace{-7mm}\text{if}\;r_{ij,i'j'}(z_{ij}, z_{i'j'}) \\ 1 - p_\phi^{ij,i'j'} & \text{otherwise}\end{cases}
\end{equation*}
where we define the relation function $r_{ij,i'j'}$ to evaluate to \textit{true} if its arguments satisfy the underlying relation, and \textit{false} otherwise; $\phi_{ijk}(z_{ij}, c_{ijk})$ is defined similarly to $\phi_{ij,i'j'}(z_{ij}, z_{i'j'})$ \footnote{We use this formulation only to accommodate settings were multiple context statements are retrieved for each query; see Section~\ref{sec:editing}. We do not have any $\phi_{ijk}$ factors if we are only using the model's predictions within a batch of test inputs as the premises for reasoning.}.
The inference problem amounts to finding $\argmax_Z \phi(Z),$ where  
\begin{equation}\label{eq:phi_z_inference}
    \phi\left(Z\right) = \prod_{i}\phi_i\prod_j\phi_{ij} \Biggl(\prod_{i'j'}\phi_{ij,i'j'}\Biggr)\left(\prod_k \phi_{ijk}\right).
\end{equation}
An approximate solution to this inference problem can be efficiently found for most problems with a MaxSAT solver such as RC2 \citep{Ignatiev2019RC2AE}. We omit arguments to the factors for conciseness. See Figure~\ref{fig:factor-graph} for a simple example of a factor graph with a single inter-belief constraint and no belief-context constraints.

\paragraph{Entailment correction.}
Consider a belief $b$, a set of its entailed statements $S=\{s_i\}_i$, unary factors $\phi(z_{b})$ and $\{\phi(z_{s_{i}})\}$, and binary factors $P=\{\phi(z_{b}, z_{s_{i}})\}_i$. Recall that an entailment relation $r_{bs_i}(z_b,z_{s_i})$ is satisfied (and the binary factor is maximized) if either $z_b=0$ or \textit{all} $z_{s_i}=1$. Consequently, as the cardinality of $\{z_{s_i} | z_{s_i} = 0\}$ increases, the more likely it is that $z_b=0$ will maximize the product of all binary factors $\prod_i \phi(z_b, z_{s_i})$. This is true even if most entailed statements are true, i.e., $|\{z_{s_i} | z_{s_i} = 1\}| \gg |\{z_{s_i} | z_{s_i} = 0\}|$. If most of the statements entailed by a belief are true, assigning the belief to be false due to a small number of (potentially spuriously) false entailed statements may be undesirable. To mitigate this outcome, we experiment with an additional type of factor in which configurations satisfying entailments with both $z_b=1$ and $z_{s_i}=1$ are `rewarded' more than other configurations satisfying the entailment:
\begin{equation*}
    \phi_{b,s_{i}}(z_{b}, z_{s_{i}}) = \begin{cases}1 & \text{if}\;z_{b},z_{s_i}=1 \\ 1 - p_\phi^{b,s_i} & \text{if}\; z_{b},z_{s_i}=0\\ 
    \sqrt{1 - p_\phi^{b,s_i}} & \text{otherwise}\end{cases}
\end{equation*}
Applying entailment correction consistently improves {\name}'s performance; see Appendix Table~\ref{tab:ec_res} for a dataset-by-dataset breakdown.

\subsection{Hyperparameters of {\name}}
\label{sec:hyperparameters}
We introduce two key hyperparameters to {\name}. Because we do not know a priori the relative reliability of the base model and relation model, we introduce the hyperparameter $\beta \in [0, 1]$, corresponding to a trade-off between the predictions of the base model and relation model. A value of $\beta=1$ corresponds to simply taking the raw predictions of the base model, while $\beta=0$ corresponds to optimizing purely for answers that are self-consistent according to the relation model, without considering the base model's beliefs. The unary factors in the factor graph become $\phi^\beta_{ij}(z_{ij}) = \left(\phi_{ij}(z_{ij})\right)^\beta$ and $\phi^\beta_{ij,i'j'}(z_{ij}, z_{i'j'}) = \left(\phi_{ij,i'j'}(z_{ij}, z_{i'j'})\right)^{1-\beta}$ (and similarly for $\phi^\beta_{ijk}$). In addition to $\beta$, we introduce a threshold $\lambda$ for relation model confidence to filter out low-confidence relation estimates. That is, we discard a relation $r_{ij,i'j'}$ or $r_{ijk}$ if $p_\phi^{ij,i'j'}<\lambda$ or $p_\phi^{ijk}<\lambda$, respectively.
In practice, we find that the optimal $\beta$ and $\lambda$ vary across problems, perhaps due to the varying complexity of the model belief and context statements (and therefore the reliability of the relation model's predictions).
Therefore, we use the \texttt{hyperopt} library \citep{bergstra2013hyperopt} for automated hyperparameter optimization, using the Tree Parzen Estimator (TPE) algorithm to tune $\beta$ and $\lambda$ jointly. We use the optimal hyperparameters found on the validation data for each problem to compute test performance. Appendix~\ref{sec:hp-tuning-appendix} details hyperparameter tuning for each experiment.


\section{Experiments}

Our experiments are broadly designed to answer the high-level question: \textit{can {\name} leverage the relational knowledge in pre-trained NLI models to produce more accurate, self-consistent system behavior, without additional data or fine-tuning?} Further, we investigate {\name}'s applicability to performing test-time \textit{model editing} \citep{Sinitsin2020Editable,mitchell2021fast}, or injection of new information, and {\name}'s sensitivity to the choice of hyperparameters and types of relations detected.


\subsection{Internal Consistency in Closed-Book Question-Answering}
\label{sec:beliefbank}


\textbf{Protocol.} To evaluate the accuracy and consistency of a set $B$ of beliefs, \citet{kassner2021beliefbank} synthesize a gold standard for those beliefs \emph{and} the inferred relations $R$. Following this prior work, we assume the following is given: 
\begin{itemize}[noitemsep,nolistsep]
    \item A set of entities $s_m \in S$
    \item A set of unary predicates $P_n \in P$
    \item A collection of ``facts'' $\left(P_n(s_m)\right)_i$, whose binary truth value is known 
    \item A directed graph of gold-standard constraints $G(P, E)$, whose edges $(P_n, P_{n'}) \in E$ represent first-order logical formulae $\forall x \left(P_n(x) \rightarrow P_{n'}(x)\right)$
\end{itemize}
From these, we construct simple yes/no questions using natural language templates. For example, for fact $P_n(s_m)$, if entity $s_m$ represents \quotestring{a lion} and predicate $P_n$ represents \quotestring{an ability to drink liquids}, the template-generated gold question answer pair $(q_i, a_i)$ is Q:~\quotestring{Is it true that a lion is able to drink liquids?}; A:~\quotestring{Yes.}  

These questions are given as input to one of two sizes of a multi-angle question answering model \citep{tafjord2021general}, given a multiple choice angle with choices \quotestring{Yes.}~and \quotestring{No.} The questions and retrieved answers $(q_i, \hat{a}_i)$ form a set of beliefs $B_{s_m}$ for each entity. Since these are closed-book questions, no context statements are supplied; because they are yes/no questions, only one candidate answer is obtained, i.e.,  $J = 1$. Question-answer to statement conversion is applied to all questions with a default answer of \quotestring{Yes.}~regardless of the answer $\hat{a}_i$, in order to provide the relation model with positive natural language assertions from which to infer sets of relations $R_{s_m}$; where the base model answers $\hat{a}_i$ are \quotestring{No.}~we replace node $z_i$ in the factor graph with its complement.  Configurations $Z_{s_m}$ are found for each $s_m \in S$ which maximize Equation \ref{eq:phi_z_inference} given $B_{s_m}, R_{s_m}$ and together form a global solution $Z$. 

\noindent \textbf{Datasets.}
\citet{kassner2021beliefbank} provide a suitable database with 12,636 facts (``silver facts''), each indicating whether one of 601 predicates relates to one of 85 entities, as well as 4,060 confidence-weighted first-order constraints manually gathered from ConceptNet \citep{conceptnet}, forming a constraint graph $G$. Additionally, they provide 1,072 distinct ``calibration facts'', each relating one of 7 entities to one of 334 predicates.  

We tune $\beta$ and $\lambda$ using a validation set of questions generated from the calibration facts, and evaluate test time performance with questions generated from silver facts.  


\begin{table}
    \setlength{\tabcolsep}{4.5pt}
    \footnotesize
    \centering
    \begin{tabular}{lcccccc}
    \toprule
        & \multicolumn{2}{c}{\textbf{Base}} & \multicolumn{2}{c}{\textbf{{\name}}} & \multicolumn{2}{c}{\textbf{G.C.}} \\
        \cmidrule(lr){2-3} \cmidrule(lr){4-5} \cmidrule(lr){6-7}
        \textbf{Model} & F1 & Con. & F1 & Con. & F1 & Con. \\
        \midrule
        Mac-Lg & 0.831 & 0.835 & 0.914 & 0.920 & 0.862 & 0.934 \\
        Mac-3B & 0.855 & 0.871 & 0.931 & 0.947 & 0.905 & 0.936 \\
    \bottomrule
    \end{tabular}
    \caption{\footnotesize F1 and consistency (1 - $\tau$) for two sizes of Macaw \citep{tafjord2021general} QA models, comparing {\name} to a naive QA baseline (Base) and {\name} with gold constraints (G.C.). {\name} significantly improves both F1 and consistency for both models.}
    \vspace{-2mm}
    \label{tab:beliefbank-results}
\end{table}

\noindent \textbf{Metrics.}
We measure accuracy using binary F1 between elements $z_{i}$ of the configuration Z maximizing $\phi(Z)$ (as in Equation~\ref{eq:phi_z_inference}), and the truth value of facts $\left(P_n(s_m)\right)_i$.  As in \citet{kassner2021beliefbank}; we use F1 for evaluation because gold answers are highly biased towards true \quotestring{No.}~answers.

We compute consistency within batches of questions using the complement of of \citet{li-etal-2019-logic}'s conditional constraint violation metric $\tau$, defined here as the proportion of \emph{relevant} gold constraints in $G$ which are \emph{violated}; a constraint $\forall x \left(P_n(x) \rightarrow P_{n'}(x)\right)$ is relevant iff, for some entity $s_m$, there is some belief $b_i \in B_{s_m}$ from fact $\left(P_n(s_m)\right)_i$ such that $z_i = 1$, \emph{and} there is some belief $b_j \in B_{s_m}$ that corresponds to fact $\left(P_{n'}(s_m)\right)_j$; the constraint is violated when $z_j = 0$.

\noindent \textbf{Comparisons.}
{\name} is evaluated against a naive baseline where only base model answers $\hat{a}_i$ and probabilities are considered. A second baseline (G.C.) performs the inference described in Sec. \ref{sec:inference}, replacing the inferred relations $R$ with the gold constraints from constraint graph $G$, rather than those estimated by the relation model.


\noindent \textbf{Results.}
Results are shown in Table~\ref{tab:beliefbank-results}. {\name} provides an absolute improvement of over 8\% in F1 and consistency for Macaw-Large and 7\% for Macaw-3B compared to the baseline. Notably, the margin of superiority of the Macaw-3B base model is mostly preserved after applying {\name}, suggesting that {\name} may provide a significant benefit even for very large models. A surprising result is that {\name} shows marked improvements in F1 over the gold constraint baseline, suggesting that the detection and filtering of relations {\name} provides may, in this setting, be an improvement over rigid adherence to the logical connections specified \emph{a priori} in \citet{kassner2021beliefbank}.


\subsection{Internal Consistency in VQA}
\label{sec:cvqa}

\begin{table}
    \setlength{\tabcolsep}{5pt}
    \footnotesize
    \centering
    \begin{tabular}{lcccccc}
    \toprule
        & \multicolumn{2}{c}{\textbf{Base}} & \multicolumn{2}{c}{\textbf{{\name}}} & \multicolumn{2}{c}{\textbf{Oracle}} \\
        \cmidrule(lr){2-3} \cmidrule(lr){4-5} \cmidrule(lr){6-7}
        \textbf{Model} & Acc. & P.C. & Acc. & P.C. & Acc. & P.C. \\
        \midrule
        LXM & 0.656 & 0.360 & 0.706 & 0.409 & 0.824 & 0.572\\
        ViLT & 0.784 & 0.489 & 0.804 & 0.548 & 0.882 & 0.690 \\
    \bottomrule
    \end{tabular}
    \caption{\footnotesize ConVQA accuracy (Acc.) and perfect consistency (P.C.) of LXMERT \citep{tan2019lxmert} and ViLT \citep{kim2021vilt} VQA models with and without {\name}. {\name} significantly improves accuracy and consistency of both models. Oracle performance is top-2 performance, as {\name} attempts to select the best of the top 2 answer choices of the base model.}
    \vspace{-6mm}
    \label{tab:vqa-results}
\end{table}
\textbf{Protocol.} The Visual Question Answering (VQA) task involves a language model generating answers to questions that are directly associated with images. VQA tests for robustness and generalizability of {\name} as it introduces an additional layer of difficulty; the task moves away from purely text-based tasks while expanding the answer space to the vocabulary of the LM being used.  The questions from the ConVQA dataset \citep{ray-etal-2019-sunny} and its associated images from the Visual Genome dataset \citep{krishnavisualgenome} provide an apt setting to assess {\name}, as the relatedness of questions for each image provide ample opportunity for model self-inconsistency.

\begin{table*}
    \footnotesize
    \centering
    \begin{tabular}{p{4cm}p{3.5cm}p{7cm}}
    \toprule
        \textbf{Input \& Gold Answer} & \textbf{Generations} & \textbf{Added context}\\
        \midrule
        \textbf{Q:} What was the first capital city of Australia?   \textbf{A:} Melbourne & \underline{Canberra}; \textbf{\color{teal}Melbourne}; Sydney; Inverell & Melbourne was the initial capital following the 1901 Federation of Australia. \\
        \midrule
        \textbf{Q:} When does the implantation of the embryo occur? \newline \textbf{A:} around 9 days after ovulation &\underline{9 to 18 days}; \textbf{\color{red}between 6 and 12 days}; after the ovulation; on the 9th week & In humans, implantation of a fertilized ovum is most likely to occur around 9 days after ovulation, however this can range between 6 and 12 days. \\
    \bottomrule
    \end{tabular}
    \caption{\footnotesize Success and failure in editing a model's behavior with {\name} by adding new information to the context. The base model's highest confidence answer is \underline{Underlined}. \textbf{Bold} shows {\name}'s output after inference; with \textbf{\color{teal}Teal, bold} showing a successful edit increasing F1 and \textbf{\color{red}Red, bold} showing an edit that reduces F1.}
    \vspace{-2mm}
    \label{tab:editing}
\end{table*}

The ConVQA dataset consists of a set of images each associated with a group of related questions about the image, such as \textit{What color is the horse?} and \textit{Is the horse brown?} for a picture of a brown horse in a stable. We evaluate {\name} with two VQA models, LXMERT \cite{tan2019lxmert} and ViLT \cite{kim2021vilt}. For each group of questions $Q_n=\{q_{ni}\}_i$, we sample the top-2 candidate outputs $\{\hat{a}_{ni1}, \hat{a}_{ni2}\}$ for each question, and use a pre-trained NLI model to infer the most likely pair-wise relations $R$ between outputs from different questions. We use the RC2 MaxSAT Solver to estimate the configuration that maximizes Equation \ref{eq:phi_z_inference}. 

\noindent \textbf{Metrics.} We report accuracy as the proportion of questions answered correctly across all groups. We infer consistency using a metric previously used in the literature for the ConVQA dataset called "perfect consistency" \citep{ray-etal-2019-sunny}. For all groups of related questions, a group is perfectly consistent if all its questions are answered correctly. Perfect consistency then reports the proportion of question groups that were perfectly consistent. While this is not a perfect measure of consistency as it excludes cases in which incorrect answers are consistent with each other, it still serves as a meaningful proxy since the dataset was designed such that any incorrect answer in a question group implies the presence of inconsistency. 

\noindent \textbf{Datasets.} We divide the ConVQA dataset into a "clean" (i.e. human verified and filtered) test set and a non-test set (train + val + test as defined by \citet{ray-etal-2019-sunny}). From the non-test set, we sample 10,000 random images equivalent to 123,746 questions to be used as our validation set for tuning our two hyperparameters. We use the clean test set -- 725 images and 6,751 questions -- to report our final results.


\begin{table}
    \footnotesize
    \centering
    \begin{tabular}{lccc}
    \toprule
        & \multicolumn{3}{c}{\textbf{F1}} \\
        \cmidrule(lr){2-4}
        \textbf{Model} & Base & {\name} & Oracle \\
        \midrule
        T5-Sm-NQ & 0.207 & 0.225 & 0.281 \\
        T5-Lg-NQ & 0.314 & 0.328 & 0.393 \\
        T5-3B-NQ & 0.332 & 0.351 & 0.423 \\
    \bottomrule
    \end{tabular}
    \caption{\footnotesize Using {\name} to inject contextual information into a model's decisions at test time. Injecting gold Natural Questions contexts consistently improves performance over the base model without requiring fine-tuning.}
    \vspace{-2mm}
    \label{tab:nq-results}
\end{table}
\noindent \textbf{Comparisons.} {\name} is compared with a naive baseline and a top-2 oracle upper bound. The naive baseline is the answer with the highest VQA model probability.  Top-2 oracle upper bound selects the correct answer if present within the top-2 predictions of the VQA model. Top-2 is appropriate given our use of the top-2 candidate outputs to generate inferences with NLI models.

\noindent \textbf{Results.} The final results for {\name}, baseline, and oracle upper bound are shown in Table~\ref{tab:vqa-results}. {\name} increases the accuracy of LXMERT and ViLT by 5\% and 2\% respectively, and the consistency of LXMERT and ViLT by 4.9\% and 5.9\% respectively. Examples in which {\name} correctly and incorrectly selects a candidate output different from the baseline output are shown in Figure~\ref{fig:vqa_good} and Figure~\ref{fig:vqa_bad}, respectively. In particular, the incorrect scenarios demonstrate several failure modes that may be in part responsible for the gap between {\name} and the oracle upper bound, suggesting further improvements of the components of {\name} will also continually improve {\name}.




\subsection{Test-Time Information Injection}
\label{sec:editing}
\textbf{Protocol.} We perform an additional experiment to evaluate {\name}'s ability to integrate external factual information into its inference process, rather than only using other predictions in the test batch. Such an ability enables editing a model's behavior at test time, without re-training, as new information becomes available. We use the Natural Questions (NQ; \citet{nq}) dataset, rather than BeliefBank, to provide more challenging inputs to the relation model. Given a question from NQ, a sentence from the ground truth context document containing information about the answer is retrieved and provided as an additional input to {\name}; we constrain the node representing this context variable in the factor graph to be true. Constraints are predicted between each answer choice and the context statement. As in the other experimental settings, hyperparameters are tuned on the validation set and applied on the test set. See Appendix~\ref{app:hyp} for tuning procedures.


\noindent \textbf{Metrics.} Model performance is evaluated using the SQuAD F1 score for overlapping tokens\footnote{\url{https://worksheets.codalab.org/bundles/0xbcd57bee090b421c982906709c8c27e1}}, following the same answer normalization protocols, including lower-casing and removing punctuation.

\noindent \textbf{Datasets.} The NQ development set consists of 7830 open-book question-answer pairs, with both long and short gold annotations in their context passages. Since the NQ test set is not available, we create a test and validation set from the NQ validation questions as follows: we take the first 5000 questions to form our test set, and the rest to be our val set, which we use for hyperparameter tuning. Then each set is filtered such that only the answerable questions remain. ``Answerable'' is defined as having a ``short answer" span defined in the annotations. This filtering process gives 2713 test entries and 1576 val entries.

\noindent \textbf{Comparisons.} {\name} is compared with a naive baseline and an oracle upper bound. All of these approaches operate on the fixed set of QA model answers for a specific QA model (one of T5-Sm-NQ, T5-Lg-NQ, and T5-3B-NQ), specifically the set of top-4 answers for each question. The naive baseline selects the answer with the highest QA model probability, $\argmax_{\hat{a}_{ij}} p_\theta(\hat{a}_{ij}|q_i)$. The oracle upper bound approach selects the answer that has the best score with the gold short answer span, $\argmax_{\hat{a}_{ij}} F_1(\hat{a}_{ij}, a_{ij})$. 

\noindent \textbf{Results.} The results on the test set using the naive baseline, {\name}, and oracle upper-bound are reported in Table~\ref{tab:nq-results}. {\name} always outperforms the naive approach, demonstrating that the framework is useful even when each query input is processed independently (i.e., non-transductively). However, despite providing a relative gain of as high as 8.7\% over the naive baseline, there is still a gap between {\name} and the oracle. This gap may be attributable to the complexity of the NQ questions and context information compared with the statements in prior experimental settings. \citet{chen2021can} demonstrate a significant gain in calibration performance from training on MultiNLI \cite{williams-etal-2018-broad} to training on a combination of MultiNLI and their NLI corpus adapted from NQ, perhaps hinting that crucial knowledge present in Natural Questions is not covered in MultiNLI, partially explaining the gap between {\name} and oracle F1 performance. Overall, these results suggest that {\name} can reason between context statements and model beliefs in addition to pairs of model beliefs, improving performance even with the increased complexity of the data.
\begin{table}
    \centering
    \small
\setlength{\tabcolsep}{4pt}
    \begin{tabular}{lcccc}
        \toprule
        & & \multicolumn{3}{c}{\textbf{F1/Accuracy}} \\
         \cmidrule(lr){3-5}
        \textbf{Model} & \textbf{Task} & {\name} & Only cont. & Only ent.  \\
        \midrule
        Mac-Lg & BB & \textbf{0.914} & 0.892 & 0.827 \\
        Mac-3B & BB & \textbf{0.931} & 0.865 & 0.917 \\
        LXM & CVQA & \textbf{0.706} & 0.691 & 0.700 \\
        ViLT & CVQA & \textbf{0.804} & 0.792 & 0.800 \\
        T5-Sm-NQ & NQ & 0.225 & 0.225 & 0.225 \\
        T5-Lg-NQ & NQ & 0.328 & \textbf{0.331} & 0.330 \\
        T5-3B-NQ & NQ & \textbf{0.351} & 0.349 & 0.350 \\
        \bottomrule
         & 
    \end{tabular}
    \vspace{-2mm}
    \caption{\footnotesize Ablating the relation types considered in {\name}'s inference procedure. The \textbf{Only cont.} and \textbf{Only ent.} are the results of applying {\name} with all entailment or contradiction relations removed, respectively. The \textbf{{\name}} column is a reproduction of the results from Sections 4.1-4.3, for convenience. Value shown is F1 score for BeliefBank (BB) and Natural Questions (NQ) and accuracy for ConVQA (CVQA). Note that hyperparameters $\beta$ and $\lambda$ are re-tuned on the respective validation set for each setting.}
    \vspace{-2mm}
    \label{tab:relation_ablation}
\end{table}

\noindent \textbf{Qualitative Analyses.} Examples of ``good'' and ``bad'' edits (edits that improve and decrease the resulting F1-scores respectively) are presented in Table~\ref{tab:editing}, with more in Appendix~\ref{app:edits}. 
When the correct answer is not available in the candidate outputs, {\name} is capable of pushing towards more partially correct answers and those that have more overlap with the context.

\subsection{Ablating Relation Types}
Given that we consider two types of relations in our experiments, contradiction and entailment, it is natural to wonder the relative contribution of these to {\name}'s performance improvement; Table~\ref{tab:relation_ablation} shows the results of this ablation. We re-run {\name} with either entailment or contradiction relations removed, re-tuning the hyperparameters for both of the new settings (contradiction-only or entailment-only). We find that the relative contribution of contradiction and entailment relations varies significantly across models even within the same task, but using both relation types always performs approximately as well or better than using just one, suggesting that both types of detected relations from the NLI model carry useful information. However, we observe in several cases, such as ViLT and the T5 models, that the entailment and contradiction relations may encode somewhat redundant information, as the performance when including either type of constraint alone nearly matches that of using both types.

\subsection{Hyperparameter Sensitivity}
We perform several experiments to clarify the relationship between the key hyperparameters, including the specific relation NLI model, $\beta$, and $\lambda$.

\begin{figure}
    \centering
    \includegraphics[width=0.9\columnwidth]{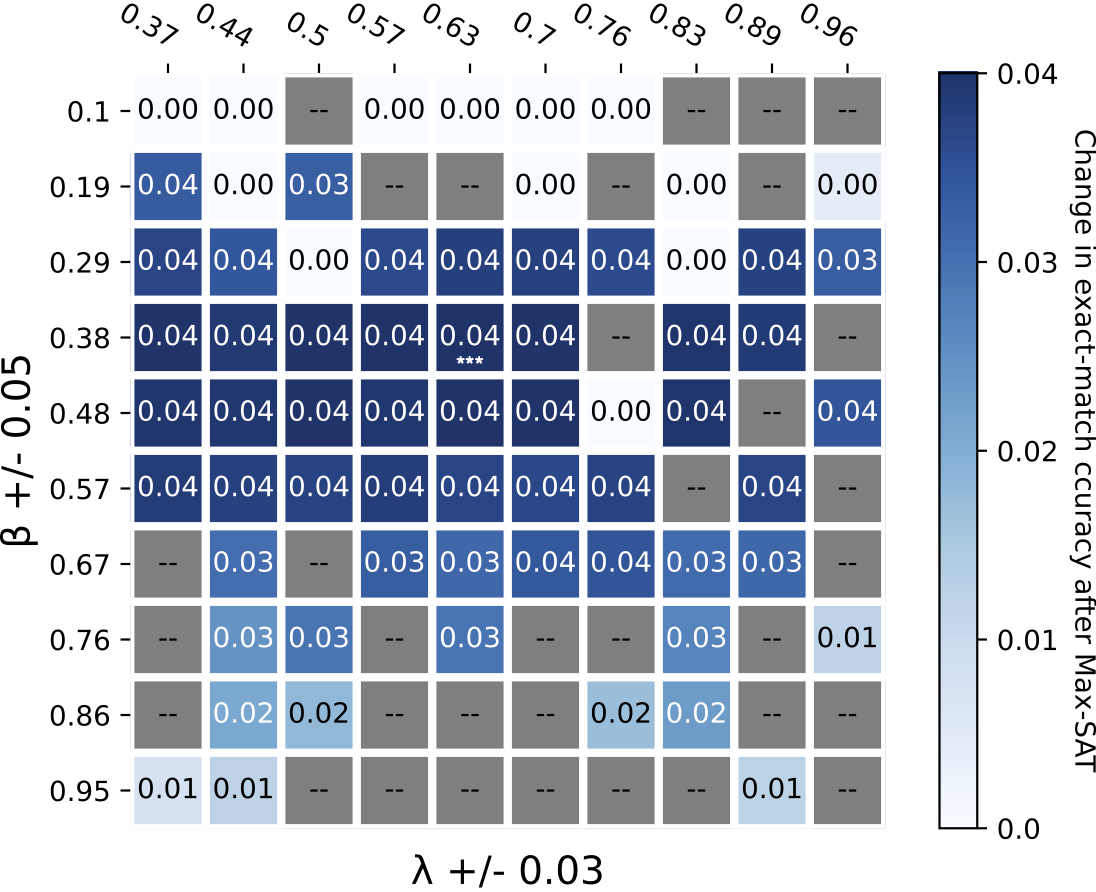}
    \caption{\footnotesize Change in {\name}'s exact-match validation accuracy as $\lambda$ (the NLI confidence threshold) and $\beta$ (tradeoff between base model and relation model beliefs) vary, holding relation model RoBERTa-Large ANLI constant. By comparing the maximum value within each column or row, we conclude that {\name} is relatively robust to the choice of $\lambda$, which the choice of $\beta$ is more important. Values are those encountered during tuning with base model ViLT on ConVQA validation questions.  Gray squares correspond to regions not evaluated during search, and asterisks (***) mark the region where the maximum increase in accuracy occurs.}
    \vspace{-2mm}
    \label{fig:beta-ablation}
\end{figure}

\paragraph{Impact of varying relation model.} Table~\ref{tab:nli-comparison} shows a comparison of {\name}'s test performance for several NLI models for each setting; notably, the best-performing NLI model is not consistent across problems. While the Albert-XXL model from \citet{nie-etal-2020-adversarial} is the strongest performing model on NQ, the simpler RoBERTa-Large models outperform it on BeliefBank and ConVQA.
\paragraph{Sensitivity to $\beta$ and $\lambda$.}
Figure~\ref{fig:beta-ablation} shows the performance of {\name} on ConVQA with ViLT as $\beta$ (the tradeoff between base model and relation model beliefs) and $\lambda$ (the NLI confidence threshold) are varied, using the values explored during hyperparameter optimization. Section \ref{sec:appendix-hyperparam-viz} of the Appendix shows similar visualizations for different VQA experiments.  If multiple hyperparameters within a grid element were explored, the best performing configuration is shown. While the maximum value in each column is the same (0.04), indicating that there \textit{exists} a good value of $\beta$ for almost any $\lambda$, the converse is not true; for some values of $\beta$, no good value of $\lambda$ exists. Thus, we conclude that the tradeoff parameter $\beta$ is the more important parameter to tune carefully.

\section{Discussion \& Conclusion}
We have presented the {\name} framework for enforcing self-consistency in pre-trained language models using relations estimated by pre-trained NLI models, showing that it improves over off-the-shelf performance in a variety of settings without requiring any fine-tuning. Our findings suggest that existing pre-trained NLI models can be a useful building block for boosting performance of NLP systems by providing useful estimates of logical relationships between model predictions across various models and datasets for QA and visual QA. 

{\name} also suggests several directions for future work. Integrating {\name} with methods that \textit{generate} questions likely to elicit useful knowledge for answering the question at hand \citep{ray-etal-2019-sunny,shwartz-etal-2020-unsupervised} may further improve performance. In addition, integrating a framework such as {\name} with recent methods for differentiation through black box combinatorial solvers \citep{Pogancic2020Differentiation} may enable training of the entire base model, relation model, and inference pipeline end-to-end, potentially further improving aggregate performance. Finally, {\name}'s general mechanism of re-ranking predictions by estimating the self-consistency of groups of model predictions is applicable beyond natural language, and future work might investigate its application to problems in vision or sequential decision-making. We hope that {\name} may serve as another promising example of integrating both neural and explicit symbolic inference machinery
into a broader intelligent system that outperforms any of its components individually. 

\begin{table}
    \footnotesize
    \centering
    \begin{tabular}{lcccc}
    \toprule
         & & \multicolumn{3}{c}{\textbf{F1/Accuracy}} \\
         \cmidrule(lr){3-5}
        \textbf{NLI Model} & \textbf{Data} & BB & ConVQA & NQ \\
        \midrule
        Alb-XXL & ANLI & 0.892 & 0.689 & \textbf{0.351}\\
        RoB-Lg & ANLI & \textbf{0.931} & \textbf{0.706} & 0.344 \\
        RoB-Lg & MNLI & 0.918 & \textbf{0.706} & 0.346 \\
    \bottomrule
    \end{tabular}
    \caption{\footnotesize Comparing {\name}'s performance for various NLI models on BB (BeliefBank), ConVQA, and NQ. Performance is measured as F1 score between predicted and gold text for BB and NQ, exact match accuracy for ConVQA. We use Macaw 3B for BB results, LXMERT for VQA results and T5-3B for NQ results. The best NLI model(s) in each column are bolded; the best NLI model varies across problems.}
    \vspace{-2mm}
    \label{tab:nli-comparison}
\end{table}

\section{Limitations}

While our results suggest {\name} can effectively leverage additional compute to boost model performance without fine-tuning, our work has some limitations. Although {\name} is conceptually applicable to generations from any language model, our work focuses on question-answering settings to leverage existing self-consistency benchmarks. In addition, {\name} increases the compute costs of inference, although it does not require fine-tuning. Further, our results suggest that the best NLI model to use for {\name} may vary across domains, requiring some tuning. As NLI models improve, we might hope that the final performance of {\name}-like systems should also inherit these gains, but Table~\ref{tab:nli-comparison} suggests that the factors that make a particular NLI model well-suited to a particular problem are not obvious, requiring further investigation.

\section*{Acknowledgements}
The authors would like to thank the anonymous reviewers for their helpful feedback during the review period, Gabe Mudel, Julie Wang, Cameron Tew, Anthony Tzen, Kevin Yang, and Ian Ng for helpful discussions and assisting with exploratory experiments early on in the project, and Nora Kassner for providing helpful early guidance in configuring the BeliefBank experiments. CF and CM are CIFAR Fellows. EM gratefully acknowledges funding from the Stanford Knight-Hennessy Graduate Fellowship. JN is supported by Stanford University Medical Scientist Training Program grants T32-GM007365 and T32-GM145402. SL acknowledges brownie bites from Target for providing a crucial fuel source for late night experiment-running.

\bibliography{citations}
\bibliographystyle{acl_natbib}

\appendix

\section{Reproducing Macaw-Large Examples}
The following configuration reproduces the Macaw-Large behavior noted in the abstract and the introduction at \url{https://huggingface.co/allenai/macaw-large}.

\noindent \texttt{\$answer\$ ;  \$question\$ = Is a sparrow a bird? ; \$mcoptions\$ = (A) Yes. (B) No. ;}

\noindent \texttt{\$answer\$ ;  \$question\$ = Does a bird have feet? ; \$mcoptions\$ = (A) Yes. (B) No. ;}

\noindent \texttt{\$answer\$ ;  \$question\$ = Does a sparrow have feet? ; \$mcoptions\$ = (A) Yes. (B) No. ;}

\section{Factor Graph Overview}
A factor graph is a factorization of a function $f$ mapping a set of $n$ variables $Z = \{z_j\}_{j=1}^n$ to a non-negative scalar. The factorization is represented as a bipartite graph containing \textit{variable nodes} and \textit{factors}; each $z_j$ is represented by one variable node, and each factor $\phi_i$ maps a subset of the variable nodes $Z_i$ to a non-negative scalar. The value of the function is computed as $f(Z) = \prod_i(Z_i)$. See \citet{loeliger2008introduction} for a more complete reference.

\section{Question-Answer to Statement Conversion Model Details}
\label{sec:qaconv-appendix}
To convert question-answer pairs into declarative statements, we combine data from the Question to Declarative Sentence (QA2D) \citep{demszky2018transforming} and BeliefBank \citep{kassner2021beliefbank} datasets to fine-tune a T5-base sequence-to-sequence model. QA2D contains question-answer pairs from five QA datasets; 95\% of the pairs are from SQuAD \citep{rajpurkar-etal-2016-squad}. The gold statements are from Amazon Mechanical Turk. The BeliefBank questions are created from silver facts using natural language templates as in Section~\ref{sec:beliefbank}, and the yes/no answers are from the known binary truth values of these facts. Our training dataset is composed of the full QA2D training dataset of 61k question-answer pairs and half of the BeliefBank silver facts, for a total of 67k training examples. Likewise, the validation dataset consists of the full QA2D validation dataset of 10k pairs and half the BeliefBank silver facts, for a total of 16k validation pairs.

The input to the QA statement conversion model is the concatenation of the question-answer pair $q_i \mathbin\Vert a_i$. Accuracy is evaluated by comparing the output sequence tokens to the gold sequence tokens. Training occurs with a learning rate of $1e^{-4}$ for a maximum of 50 steps, where each step consists of 32k training examples, with early stopping if validation loss does not decrease for 6 consecutive steps. We ran the fine-tuning on NVIDIA GeForce RTX 3090 GPU. Fine-tuning ended after 14 steps with a final training accuracy of 0.764 and validation accuracy of 0.628. This took approximately 40 minutes. Table~\ref{tab:statement-conversion-examples} demonstrates the model's performance on a few validation examples.

\begin{table*}[!ht]
    \centering
    \small
    \begin{tabular}{p{1.5cm}p{4cm}p{4cm}p{4cm}}
    \toprule
          \textbf{Dataset} & \textbf{Input} & \textbf{Output} & \textbf{Gold statement}\\
        \midrule
        SQuAD & Who established Yale's residential college system?	Edward S. Harkness & Edward S. Harkness established Yale's residential college system. & Edward S. Harkness established Yale's residential college system.\\
        SQuAD & How did Kuhn view the history of science? competing paradigms or conceptual systems & Kuhn viewed the history of science as \textbf{\color{red}a competing paradigm or conceptual system}. & Kuhn viewed the history of science as \textbf{\color{teal}competing paradigms or conceptual systems}.\\
        BeliefBank & Is it true that a poodle is a river? No & A poodle is not a river. & A poodle is not a river.\\
        BeliefBank & Is a pigeon a living thing? Yes & A pigeon is a living thing. & A pigeon is a living thing.\\
    \bottomrule
    \end{tabular}
    \caption{\footnotesize The QA statement conversion model outputs declarative statements from question-answer pairs. Out of the four validation examples presented, three are correct. The \textbf{\color{red}Red, bolded} portion of the output of the second example indicates how it differs from the \textbf{\color{teal}Teal, bolded} corresponding portion of the gold statement.}
    \label{tab:statement-conversion-examples}
\end{table*}

\begin{table}[!ht]
    \setlength{\tabcolsep}{1.5pt}
    \centering
    \small
    \begin{tabular}{cccc}
    \toprule
    & \multicolumn{3}{c}{\textbf{F1/Accuracy}} \\
    \cmidrule(lr){2-4}
       \textbf{Model} & \textbf{Naive} & \textbf{w. E.C.} & \textbf{w/o. E.C.} \\
       \midrule
        Mac-Lg+Rob/ANLI & 0.831 & \textbf{0.914} & 0.909  \\
        Mac-3B+Rob/ANLI & 0.855 & \textbf{0.931} & 0.886  \\
        LXMERT+Rob/MNLI & 0.656 & \textbf{0.706} & 0.701 \\
        LXMERT+Rob/ANLI & 0.656 & \textbf{0.706} & 0.693 \\
        ViLT+Rob/MNLI & 0.784 & 0.804 & \textbf{0.810} \\
        ViLT+Rob/ANLI & 0.784 & \textbf{0.814} & 0.807 \\
        \bottomrule
    \end{tabular}
    \caption{\footnotesize Comparison of {\name} test performance vs.~baseline with and without entailment correction (E.C.) across base+relation models for closed-book question answering (Macaw) and VQA (LXMERT, ViLT) experiments (F1 for closed-book QA, exact-match accuracy for VQA), showing that the entailment correction improves performance for most configurations.}
    \vspace{-2mm}
    \label{tab:ec_res}
\end{table}

\section{Additional Modifications to {\name}}
A timeout for solvers is imposed in order to prevent the RC2 MaxSAT solver from running optimization indefinitely. The average solve time per question was <4 ms for closed-book QA, <1 ms for VQA and <20 ms for NQ (for NQ, the solve time is < 1/10th of the time needed for a forward pass through the QA and NLI models). We found only one batch of test questions for the closed-book QA task and VQA task where the solver couldn't find a solution efficiently, so we set a short timeout (30s for CBQA, 10s for VQA, none required for NQ).

We also de-duplicate the list of inferred constraints before passing the statement and constraint groups through the MaxSAT solver so that only the highest-weighted constraints would remain among their duplicates. 

\section{Entailment Correction Ablations} \label{app:ecc}
Table \ref{tab:ec_res} shows the effects of entailment correction on {\name} test performance in closed-book question answering and VQA experiments for different choices of base model, using the NLI relation model resulting in the best test set performance (RoBERTa-Large-MNLI).

\section{Additional ``Good'' and ``Bad'' Edit Pairs} \label{app:edits}
\begin{figure}
    \centering
    \includegraphics[width=\linewidth]{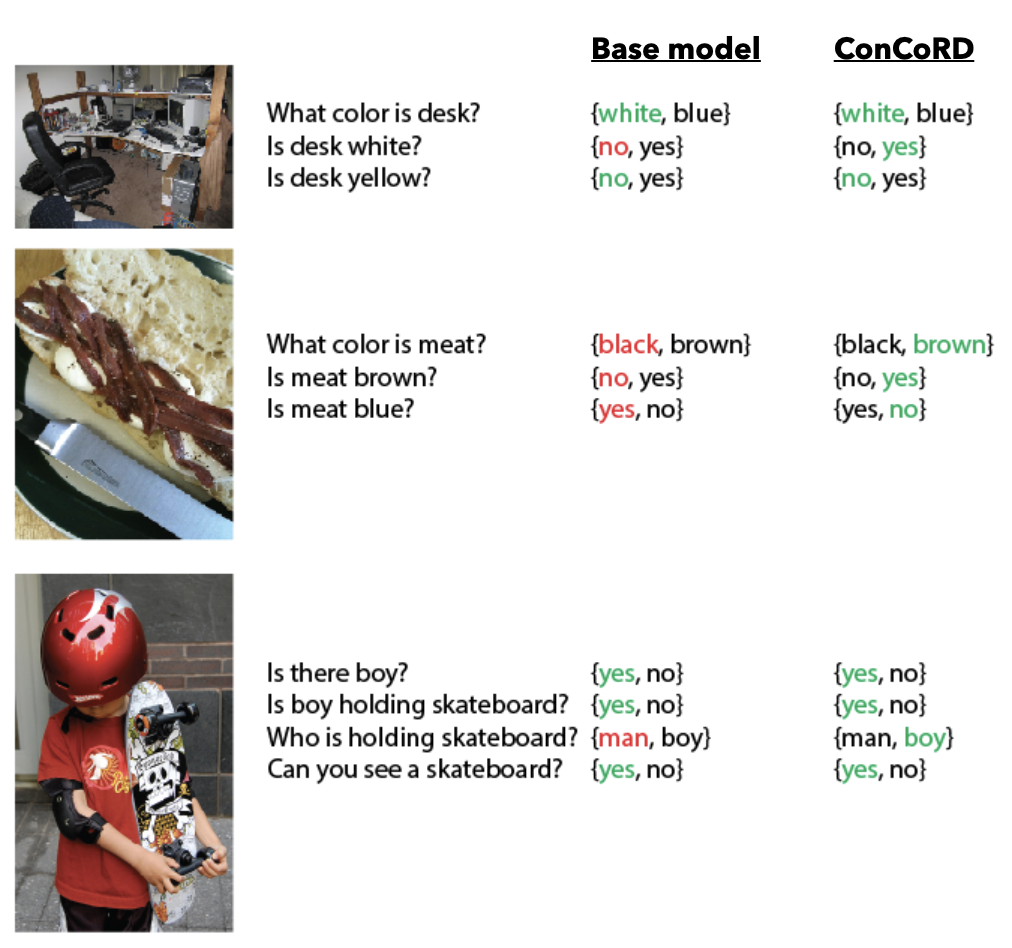}
    \caption{\footnotesize ``Good'' flip examples from the VQA experiments. The green texts mark the correctly selected answers, while the red texts indicate incorrectly selected answers.}
    \vspace{-2mm}
    \label{fig:vqa_good}
\end{figure}

\begin{figure}
    \centering
    \includegraphics[width=\linewidth]{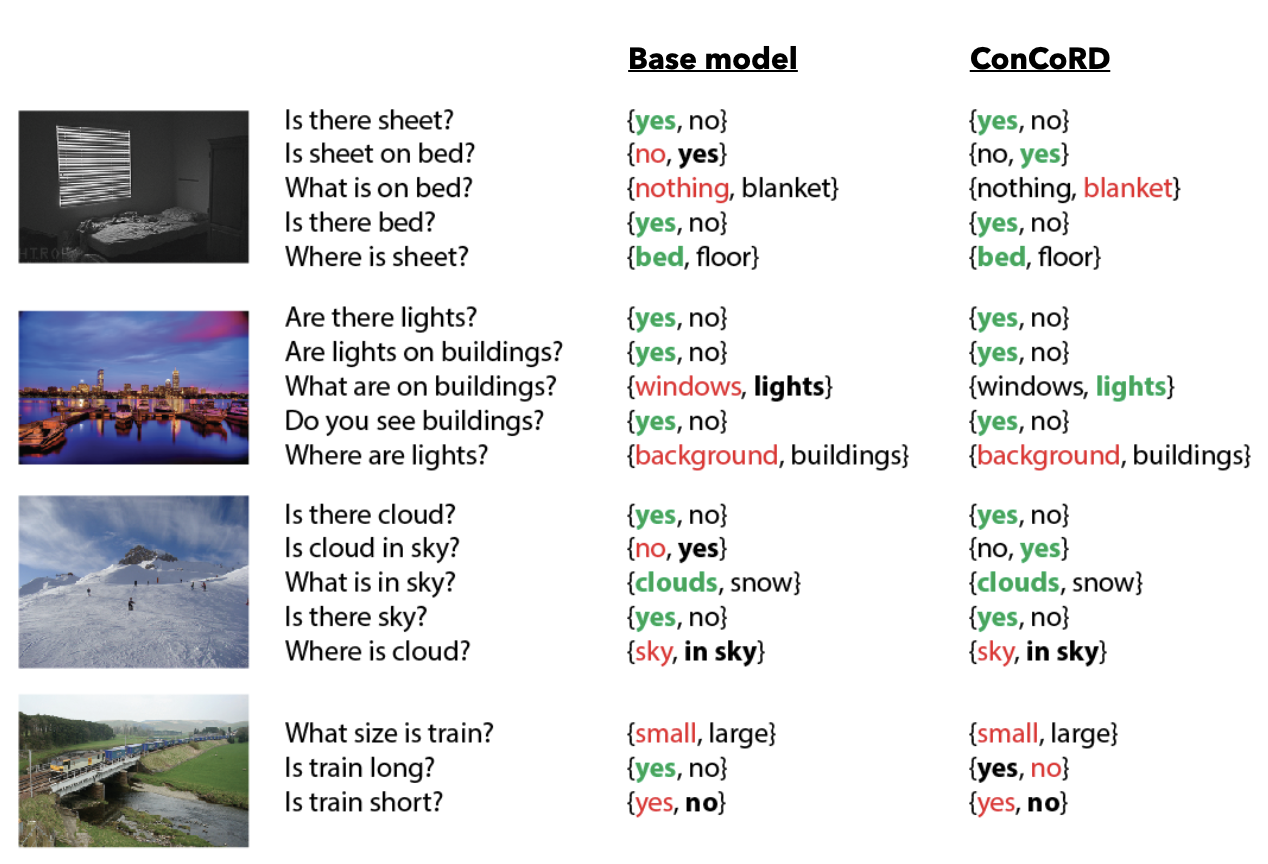}
    \caption{\footnotesize ``Bad'' flip examples from the VQA experiments. The green texts mark the correctly selected answers, while the red texts indicate the incorrectly selected answers. The \textbf{bolded} texts are the correct answers, if generated within the top-2 predictions. From top to bottom, the first image is an example of when the correct answer, "sheet," was not contained in the predicted answers. The second image is an example of when the conversion of QA pair to statement did not occur as intended and the NLI failed to generate the appropriate inferences that could be used to inform correction of "background" to "buildings. The third image shows an example of when an "incorrect" answer (sky) is effectively the same as the "correct" answer (in sky)--only semantically different. The fourth image shows an example of when the model strongly believed in an incorrect answer and changed another correct answer.}
    \vspace{-2mm}
    \label{fig:vqa_bad}
\end{figure}

More examples of good and bad edits in the Editing experiment are presented in Table~\ref{tab:editing-extra}. We also include good  (Figure~\ref{fig:vqa_good})and bad flip (Figure~\ref{fig:vqa_bad}) examples from the VQA dataset. For the bad flip examples in VQA, we include different failure modes to demonstrate the types of potential {\name} errors.

\section{Good and Bad Flips}

For each set of experiments on the test set, we report the numbers of good and bad flips made by {\name} in Table~\ref{tab:gb_flips}. It can be observed that the number of good flips is consistently significantly higher than that of bad flips.
\begin{table}[h]
    \centering
    \small
    \begin{tabular}{p{1.55cm}p{1.5cm}p{1.5cm}p{1.45cm}}
    \toprule
       \textbf{Experiment} & \textbf{Model} & \textbf{Good Flips} & \textbf{Bad Flips}\\
       \midrule
        BeliefBank & Macaw-3B & 723 & 277\\
        VQA & LXMERT & 576 & 238\\
        NQ & T5-3B-NQ & 168 & 69 \\
        \bottomrule
    \end{tabular}
    \caption{\footnotesize The numbers of good and bad flips in each of the experiments performed. We define flips as choosing a different candidate from the naive baseline for the multiple choice experiments, and a binary truth value flip for BeliefBank. ``Good'' flips are flips that improves performance, and ``bad'' flips are those that are detrimental to performance.}
    \vspace{-2mm}
    \label{tab:gb_flips}
\end{table}

\begin{table*}[!ht]
    \centering
    \small
    \begin{tabular}{p{2cm}p{3.5cm}p{3.5cm}p{5cm}}
    \toprule
          \textbf{Model} & \textbf{Input \& Gold Answer} & \textbf{Generations} & \textbf{Added context}\\
        \midrule
        T5-Sm-NQ & \textbf{Q:} Who was the declaration of independence written for? \newline \textbf{A:} the Second Continental Congress & \underline{Second Continental} \underline{Congress}; \textbf{\color{red}the United States}; the British Crown; Great Britain & The United States Declaration of Independence is the statement adopted by the Second Continental Congress 
        meeting at the Pennsylvania State House ( Independence Hall ) in Philadelphia on July 4 , 1776 , which announced that the thirteen American colonies , then at war with the Kingdom of Great Britain , regarded themselves as thirteen independent sovereign states , 
        no longer under British rule .\\
        T5-Sm-NQ & \textbf{Q:} What is the scientific name for the calf muscle \newline \textbf{A:} gastrocnemius muscle & \underline{The serratus calf muscle}; \textbf{\color{teal}gastrocnemius muscle}; The serratus calfi; The serratus muscle & Along with the soleus muscle , the gastrocnemius forms half of the calf muscle . \\
        \midrule
        T5-3B-NQ & \textbf{Q:} Who is the actor that plays Dr. Sean Murphy \newline \textbf{A:} Freddie Highmore & \underline{Freddie Highmore}; \textbf{\color{red}Daryl ``Chill'' Mitchell}; Dylan Christopher Minnette; Javier Muoz  & The series stars Freddie Highmore as Shaun Murphy , a young surgical resident with autism and savant syndrome at San Jose St. Bonaventure Hospital ., Freddie Highmore as Shaun Murphy : A surgical resident with autism and savant syndrome .\\
        T5-3B-NQ & \textbf{Q:} Who is the founder of the Ubuntu project \newline \textbf{A:} Mark Richard Shuttleworth& \underline{Linus Torvalds}; \textbf{\color{teal}Mark Shuttleworth}; Richard St. John Hopper; Richard St. John Redmond & Mark Richard Shuttleworth ( born 18 September 1973 ) is a South African entrepreneur who is the founder and CEO of Canonical Ltd. , the company behind the development of the Linux - based Ubuntu operating system .\\
    \bottomrule
    \end{tabular}
    \caption{\footnotesize Editing a model's behavior by adding new information to the context. The \underline{Underlined} generation is the answer with the highest QA model confidence. The \textbf{Bolded} generation is what {\name} selects after NLI inference. \textbf{\color{teal}Teal, bolded} generations indicate that {\name} selects a generation with higher token overlap F1, while \textbf{\color{red}Red, bolded} generations indicate that {\name} selects a worse generation.}
    \label{tab:editing-extra}
\end{table*}

\section{Hyperparameter Search Details} \label{app:hyp}
\subsection{Experiments}
\label{sec:hp-tuning-appendix}
\subsubsection{Closed-Book Question Answering}
Hyperparameters (Section~\ref{sec:hyperparameters}) are tuned jointly using \texttt{hyperopt} on the BeliefBank calibration dataset (Section~\ref{sec:beliefbank}). The search space of $\beta$ is uniform between $[0.05, 1.0]$, and for $\lambda$ it is uniform between $[0.5, 1.0]$. \texttt{hyperopt} optimizes cumulative F1 across all entity batches for 300 trials.
To speed-up tuning, we created caches of model beliefs $B_{s_m}$ and relation sets $R_{s_m}$ for each calibration entity $s_m$. This was run on NVIDIA GeForce RTX 3090 GPU, and the largest NLI models took up to two hours to complete. Using these caches, \texttt{hyperopt} tuning completes in less than an hour on CPU. The best performance on the calibration facts for each of the base Macaw models is reported in Table~\ref{tab:beliefbank_val_results}. The results show that $\beta$ is higher for the better base model Macaw-3B. 

\begin{table}[!h]
    \centering
    \small
    \begin{tabular}{ccccccc}
    \toprule
        \textbf{Model} &  \textbf{F1} & $\beta$ & $\lambda$ & \textbf{E.C.}  \\
        \midrule
       Macaw-Large  & 0.919 & 0.753 & 0.855 & True\\
       Macaw-3B & 0.94 & 0.804 & 0.873 & True\\
       \bottomrule
    \end{tabular}
    \caption{Validation performance on the BeliefBank calibration facts. Both models achieve best validation performance with the RoBERTa-Large ANLI model.}
    \vspace{-2mm}
    \label{tab:beliefbank_val_results}
\end{table}

\subsubsection{VQA}

Hyperparameters are tuned jointly using \texttt{hyperopt}.  The search space for $\beta$ is uniform over $[0.05, 1]$, for $\lambda$ it is uniform over $[\frac{1}{3}, 1]$.  A total of 100 trials were performed, updating parameters using TPE, on an AWS \texttt{g4dn.xlarge} EC2 instance.  Each search took less than one hour. Table \ref{tab:vqa_hyperparams_val} shows the selected parameters and their exact-match accuracy on validation questions.

\begin{table}[!h]
    \centering
    \small
    \begin{tabular}{ccccccc}
    \toprule
        \textbf{VQA} &  \textbf{Acc.} & $\beta$ & $\lambda$ & \textbf{E.C.}  \\
        \midrule
       LXMERT  & 0.691 & 0.208 & 0.805 & True\\
       ViLT & 0.787 & 0.395 & 0.772 & True\\
       \bottomrule
    \end{tabular}
    \caption{Validation performance on VQA. Both models achieve best validation performance with the RoBERTa-Large MNLI model.}
    \vspace{-2mm}
    \label{tab:vqa_hyperparams_val}
\end{table}

\subsubsection{Information Injection with Natural Questions}
For this round of experiments, we lower the bounds for $\beta$ and $\lambda$ after some initial trials. The bounds of $\beta$ are $[0, 0.5]$ and the bounds of $\lambda$ are $[0, 0.6]$. We run \texttt{hyperopt} for 200 trials (often taking approximately 2 to 3 hours on an NVIDIA GeForce RTX 3090 GPU) for each of the three NLI models. \texttt{Hyperopt} optimizes for the highest token-overlapping F1 score in this experiment. 

We report the best validation performance of each of the QA base models in Table~\ref{tab:nq_validation_performance}.

\begin{table}[!h]
    \centering
    \small
    \begin{tabular}{cccccc}
    \toprule
        \textbf{Model} & \textbf{F1} & $\beta$ & $\lambda$ & \textbf{E.C.}  \\
        \midrule
       T5-Small & 0.227 & 0.112& 0.540 & True\\
       T5-Large & 0.331 & 0.081 & 0.413 & False\\
       T5-3B & 0.353 & 0.072 & 0.477 & True\\
       \bottomrule
    \end{tabular}
    \caption{Validation performance on NQ. All models achieve best validation performance with the ALBERT ANLI model.}
    \vspace{-2mm}
    \label{tab:nq_validation_performance}
\end{table}




\subsection{Visualizing Hyperparameter Search}
\label{sec:appendix-hyperparam-viz}

\begin{figure}[h]
    \subfloat[Base model LXMERT, with entailment correction.]{
      \includegraphics[clip,width=\columnwidth]{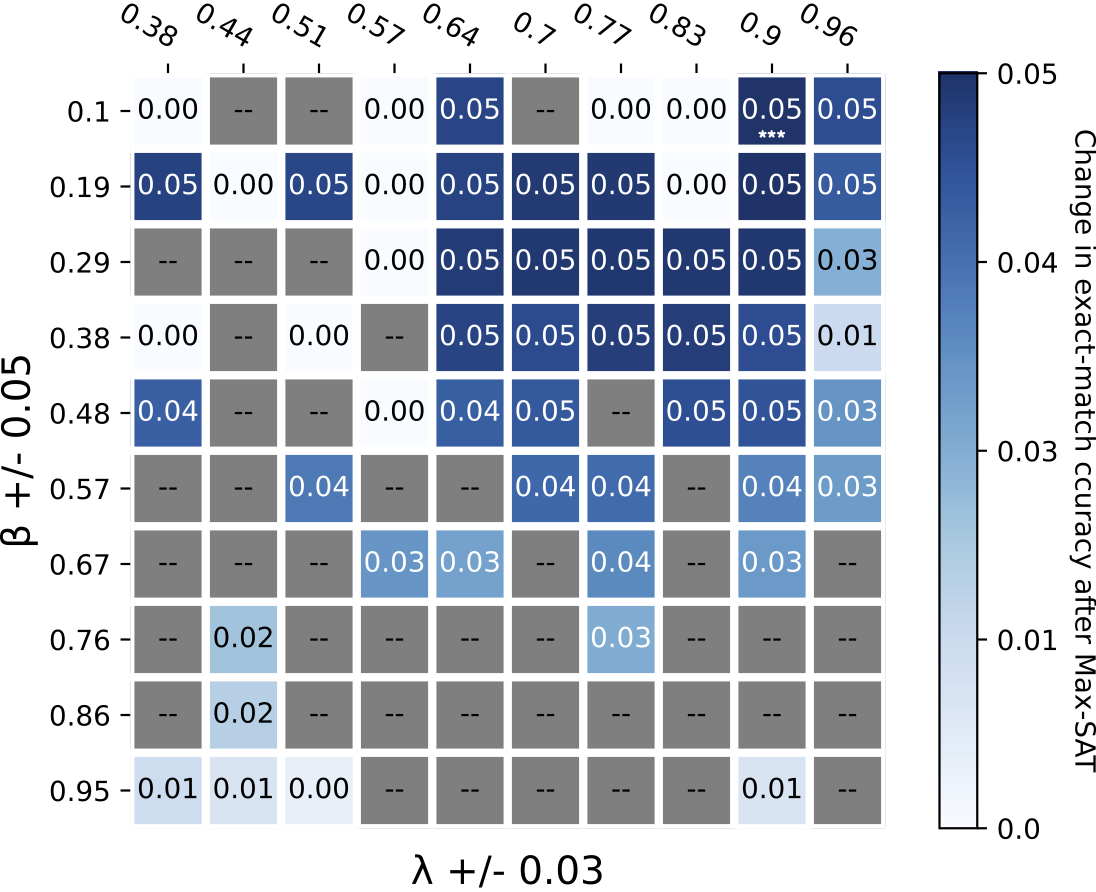}
      \label{fig:rest-a}
    } \\
    \subfloat[Base model LXMERT, without entailment correction.]{
      \includegraphics[clip,width=\columnwidth]{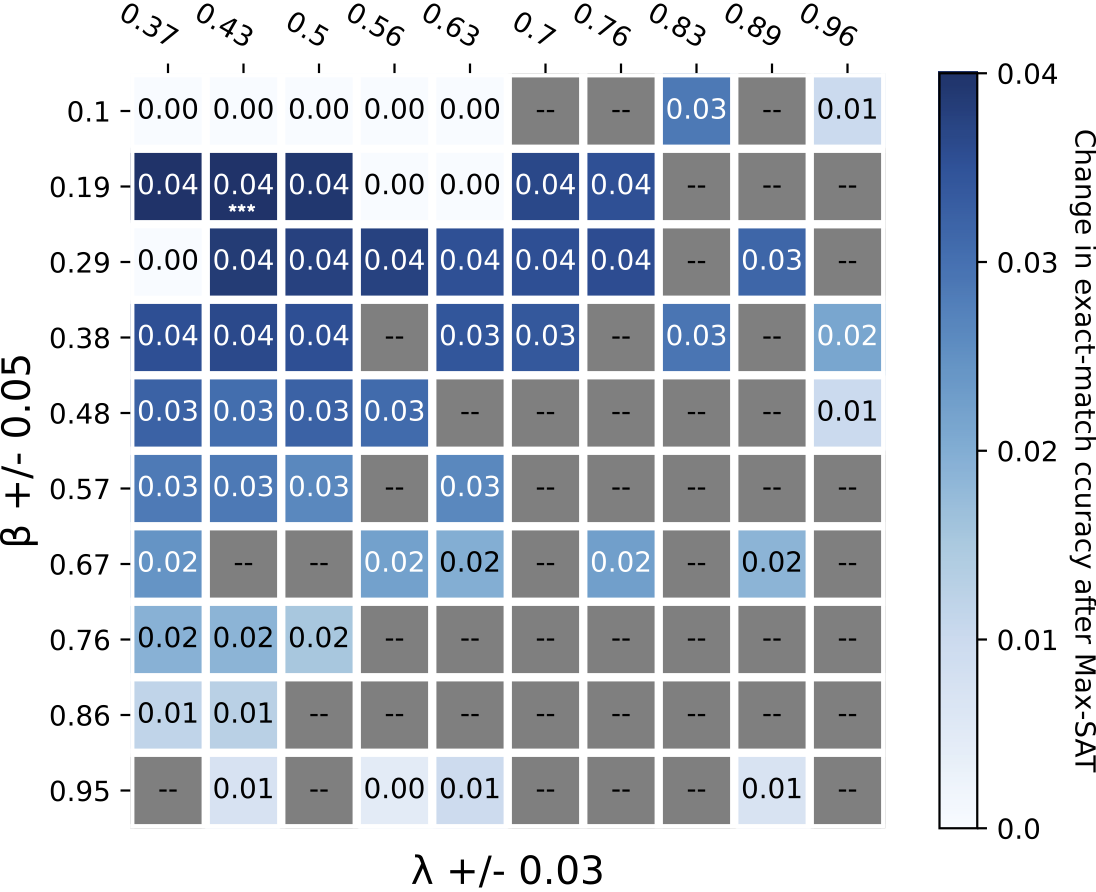}
      \label{fig:rest-b}
    } \\
    \subfloat[Base model ViLT, without entailment correction.]{
      \includegraphics[clip,width=\columnwidth]{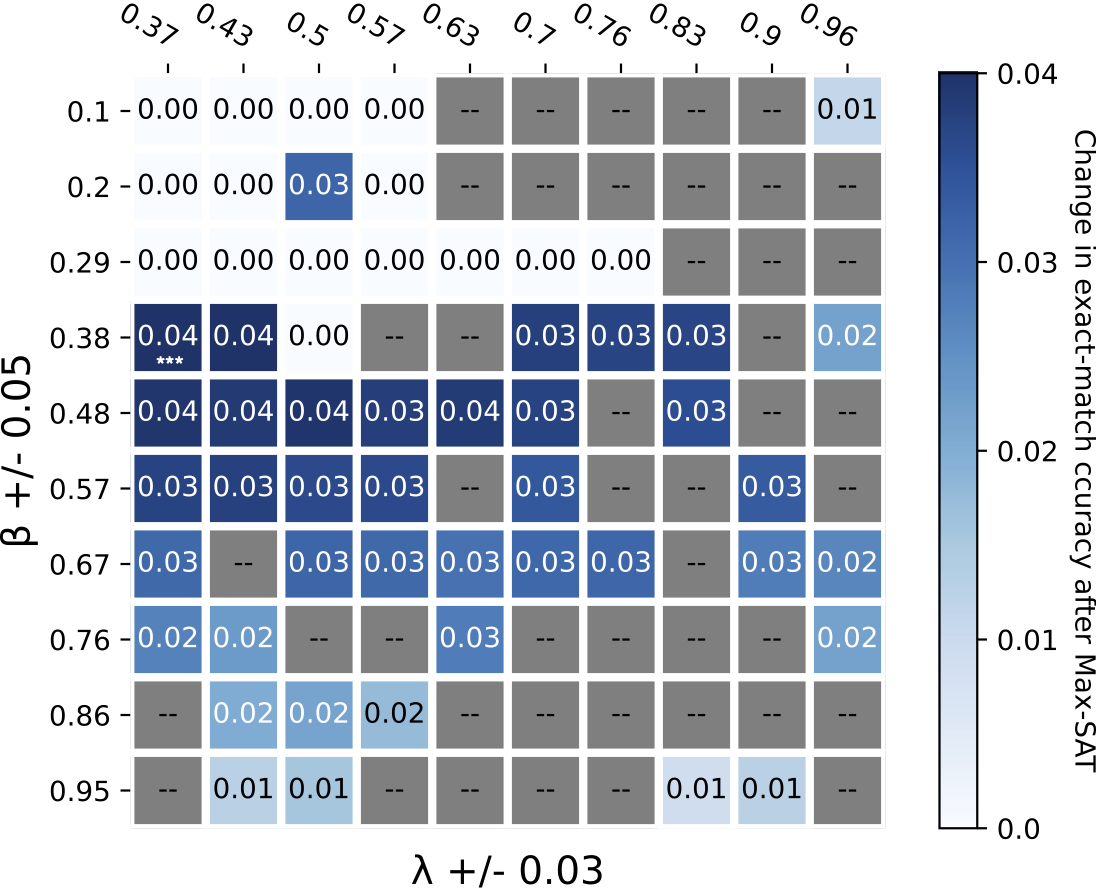}
      \label{fig:rest-c}
    } \\
    \caption{\footnotesize 
        As in figure \ref{fig:beta-ablation}, we show changes in exact-match validation accuracy as a function of confidence threshold $\lambda$ and tradeoff parameter $\beta$, with several choices of base model, with and without an entailment correction, holding relation model RoBERTa-Large ANLI constant.  
    }
    \label{fig:rest-of-heatmaps}
\end{figure}

Figure \ref{fig:rest-of-heatmaps} shows increases in exact-match accuracy as they vary with choices of $\lambda$, $\beta$, for additional choices of base model for a VQA task, with and without entailment correction, complementing figure \ref{fig:beta-ablation}.  Interestingly, choosing a different base model does noticeably effect the optimum value of $\beta$; between figures \ref{fig:rest-b} and \ref{fig:rest-c} we see the near-optimal region shift towards a value of $\beta$ that gives higher confidence in the base model where the base model produces ``better'' answers.  However, the increase in accuracy is similar, suggesting that with appropriate selection of $\beta$, {\name} can offer similar improvements over a range of choices of base model. 

\end{document}